\documentclass{article} 
\usepackage{iclr2021_conference,times}

\usepackage{amsmath,amsfonts,bm}

\def\eqref#1{equation~\ref{#1}}

\def\1{\bm{1}}

\DeclareMathAlphabet{\mathsfit}{\encodingdefault}{\sfdefault}{m}{sl}
\SetMathAlphabet{\mathsfit}{bold}{\encodingdefault}{\sfdefault}{bx}{n}

\usepackage[hidelinks]{hyperref}
\usepackage{url}
\usepackage{todonotes}

\usepackage{graphicx}
\graphicspath{ {.} }

\title{Probing artificial neural networks: \\ insights from neuroscience}

\author{Anna A.~Ivanova \\
MIT\\
\texttt{annaiv@mit.edu} \\
\And
John Hewitt \\
Stanford University \\
\texttt{johnhew@stanford.edu} \\
\And
Noga Zaslavsky \\
MIT\\
\texttt{nogazs@mit.edu} \\
}

\iclrfinalcopy 
\begin{document}

\maketitle

\begin{abstract}
A major challenge in both neuroscience and machine learning is the development of useful tools for understanding complex information processing systems. One such tool is probes, i.e., supervised models that relate features of interest to activation patterns arising in biological or artificial neural networks. Neuroscience has paved the way in using such models through numerous studies conducted in recent decades.
In this work, we draw insights from neuroscience to help guide probing research in machine learning.
We highlight two important design choices for probes --- direction and expressivity --- and relate these choices to research goals. We argue that specific research goals play a paramount role when designing a probe and encourage future probing studies to be explicit in stating these goals.
\end{abstract}

\section{Introduction}

Human brains and contemporary artificial neural networks (ANNs) share a fundamental property: both are complex information processing systems that can perform hard cognitive tasks, such as object recognition and natural language inference. The complexity of these systems poses a major challenge for understanding the internal computations and representational structures that enable their success, leading to a large body of research that aims to improve the interpretability of brains and ANNs.

Here, we focus on one important tool that has been employed in both fields, known in the ML literature as \emph{probing}. 
The idea of training a model to `probe' internal representations of a computational system originated in neuroscience \citep[e.g.,][]{cox2003functional, kamitani2005decoding, mitchell2008predicting} and was later introduced to study ANNs by several different research groups \citep{ettinger2016probing, adi2016fine, gelderloos2016phonemes,alain2016understanding}. In these works, a classifier is trained to predict a set of target features from activations evoked by a set of stimuli (e.g., sentences) in an ANN layer or a brain region. If the classifier performs well, it suggests that the (artificial or biological) neural system is informative of the target features, and therefore these features can potentially be used in downstream computations.

Recently, the machine learning (ML) community has been debating the usefulness of probes and discussing best practices for probe design and interpretation \citep[e.g., ][]{pimentel2020information, hewitt-liang-2019-designing, belinkov2021probing, elazar2021amnesic}. We aim to contribute to this discussion by drawing insights from neuroscience that may inform the design of probes for ANNs.

\section{Probe design}

We define a \emph{probe} as follows. Given a set of system activations $X$ in response to a set of stimuli $I$, we want to find a mapping $F$ that relates $X$ to a set of target features $Y$. These features are defined independently from the system. A \emph{probe} is a supervised model that is trained to find the best mapping between $X$ and $Y$, subject to constraints on the functional form of $F$.

Next, we discuss two probe design choices: direction and expressivity.

\subsection{Direction}
\label{decoders}

The vast majority of ANN probes to date have been set up as decoders, trained to find a mapping $F:~X \rightarrow~Y$. However, an alternative approach is to train an encoder, defined by a mapping $F:~Y \rightarrow~X$. The tight relationship between encoders and decoders is widely recognized in neuroscience \citep{holdgraf2017encoding, naselaris2011encoding, king2018encoding}. Although similar in their setup, decoders and encoders provide different insights. Decoders allow us to ask whether features $Y$ are predictable from activations $X$ while ignoring all other sources of variance. In contrast, encoders allow us to ask how much variance in $X$ is explained by a given feature set $Y$ (where, for ANNs, $X$ can be defined as either the entire layer or an individual unit). Thus, decoders and encoders can be used hand in hand as complementary sources of knowledge about the inner workings of a system.

\subsection{Expressivity}
\label{expressivity}

An important decision when picking a probe is to decide how expressive it should be, i.e., what is the space of possible mappings for $F$. Both neuroscientists \citep[e.g.,][]{kamitani2005decoding} and ML researchers \citep[e.g.,][]{pimentel2020information} noted that unboundedly expressive probes make it difficult to draw informative conclusions from the performance of a probe. If the network layers form a Markov chain (which is true for many ANNs and is potentially true for perceptual brain regions), a feature decodable at a later processing stage will also be decodable at an earlier processing stage \citep[due to the data processing inequality; ][]{shannon1948mathematical}. 
In neuroscience, this issue is sometimes called ``decoding from the retina'': even if high-level objects are decodable from the retina, it doesn't mean that the retina itself performs object recognition.

In addition, unboundedly expressive probes provide no guarantees that the system can actually use features $Y$ when performing the task. In neuroscience, \citet{ritchie2019decoding} argue that successful decoder performance can often be explained by data artifacts rather than actual neural activity. In NLP, \citet{hewitt-liang-2019-designing} showed that an overly expressive probe can memorize random data labels, thus not revealing much about the internal feature representations within a layer.

Overall, both communities agree that decodability alone is often insufficient for understanding the role of target features within the system. 
Few principled solutions have been proposed for determining optimal probe expressivity. \citet{pimentel2020pareto} have attempted to determine the preferred probe class by considering a complexity-accuracy tradeoff; however, their approach does not account for the plurality of research goals, each of which might place unique constraints on the probe. We aim to close this gap by explicitly relating probe design choices and research goals.

\section{The role of research goals}
\label{questions}

\begin{figure}[t]
\includegraphics[width=14.5cm]{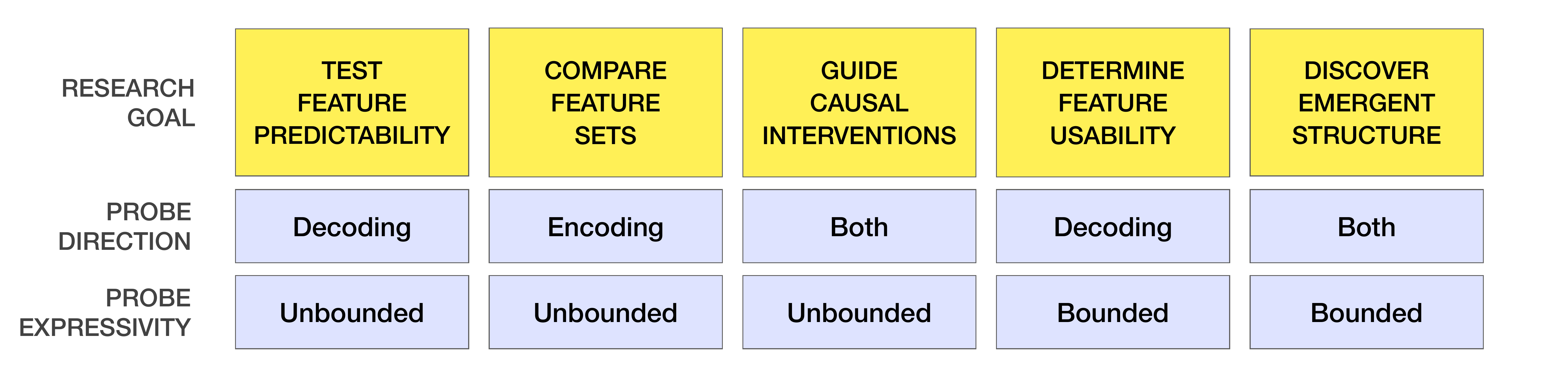}
\caption{Possible probing goals and recommended probe structure for each goal.}
\label{figg}
\end{figure}

We argue that the design of probes, including their directionality and expressivity, should be guided by specific scientific goals. 
Probing efforts often have the overarching goal, explicit or implicit, of eventually constructing improved ML systems, for example, by identifying what properties models lack \citep{liu2019linguistic}.
However, probing experiments rarely lead directly to modeling improvements; instead, they contribute to our understanding of the emergent properties of ANN representations and their relationships to better-understood features.
In contributing to this body of knowledge, each probing study implicitly (or explicitly) pursues a finer-grained research goal. The choice of a goal then helps determine the design of the probing study (Figure \ref{figg}).

Below, we list a representative set of fine-grained probing goals that, we believe, would be particularly relevant to ANN researchers. Our list is partially based on a related work from neuroscience \citep{ivanova2021simple} although, of course, only some of the goals are shared between the two communities.

\subsection{Testing feature predictability}
\label{predictability}
A basic question one can ask is whether there exists any (potentially unboundedly complex) mapping $F$ that allows us to predict features $Y$ from neural activity $X$. Common examples in neuroscience include using brain data to predict clinical disorders \citep[e.g., ][]{kazemi2018deep} or behavioral traits \citep[e.g., ][]{shen2017using}. In NLP, some studies have attempted to decode from language-only ANNs properties of the world, like the size of objects \citep{zhang2020language}. Studies in this category focus on the question ``can we decode $Y$?'' without aiming to determine how $Y$ is encoded. 

The problem formulation naturally makes it suitable for a decoder. Such a decoder can have unbounded expressivity: in order to maximize the probe's predictive accuracy, we shouldn't place a limit on the space of possible mappings. 
However, this lack of restrictions automatically raises the issues discussed in section \ref{expressivity}:~even if a probe finds a predictive mapping from $X$ to $Y$, there is no guarantee that $Y$ is used (or even usable) in downstream computations. For instance, \citet{ravichander2020probing} showed that probes trained on a synthetic dataset could decode a property that was randomly injected in half of the inputs, suggesting that decodability is not always indicative of task relevance. Further, this research question does not allow us to characterize computations at different ANN layers because, with a powerful enough probe, the features extractable from one layer will also be extractable from the rest \citep[assuming each $i$ corresponds to a unique $x$ at every layer; ][]{pimentel2020information}. Thus, feature predictability is often insufficient for elucidating the computational properties of a system of interest.
    
\subsection{Comparing feature sets}
Instead of testing the predictability of a particular feature set, one might ask which of two (or more) feature sets corresponds better to the internal representations of that system. Examples from neuroscience include comparing syntactic, semantic, and discourse-level text features  \citep{wehbe2014simultaneously} or competing syntactic parser models \citep{brennan2016abstract} in their ability to predict brain activity during language comprehension. Some NLP researchers have also used probes to adjudicate between competing syntactic formalisms \citep[e.g., ][]{kulmizev2020neural}.

For comparisons across feature sets, an encoder is arguably more useful than a decoder: the two feature sets might be equally decodable, but if one explains more variance in the system, it is likely closer to the `true' underlying representation \citep{naselaris2011encoding}. For instance, if the mapping $Y_{phonetic} \rightarrow X$ has a higher predictive accuracy than $Y_{syntactic} \rightarrow X$, we can conclude that activation in $X$ is primarily driven by $Y_{phonetic}$. Therefore, $X$ is closer to a phonetic processor than to a syntactic processor. 

If there are no additional theoretical commitments, feature set comparison studies might also choose to employ an unboundedly expressive probe. However, they would need to address the same concerns as those arising for goal \ref{predictability}. Furthermore, care must be taken to determine the commensurability of the feature sets being compared. If one feature set is substantially more complex than the other, it will likely provide a better fit to the data. In an extreme example, a feature set containing every letter in every word of a sentence will almost certainly achieve better predictivity than the feature set containing just syntactic labels, even though the former feature set provides little insight into the underlying ANN representations. Finally, if there is a deterministic mapping from one feature set to another, using an unboundedly expressive probe to compare them provides no insight whatsoever.

\subsection{Guiding causal interventions}
After establishing feature correlates within a system of interest, we often want to know whether these features are actually used by the system. Although probing is a correlational method and thus cannot determine the causal effect of feature correlates on ANN's behavior, it can serve as an important complement to causal studies. Researchers can leverage the ability of probes to identify ANN activity patterns that are linked to a given feature set, and then eliminate \citep{elazar2021amnesic} or alter \citep{giulianelli2018under} those patterns to determine whether they affect the system's behavior in expected ways. Neuroscience has a rich tradition of combining correlational and causal studies \citep[e.g., ][]{bergmann2016combining, caldwell2019direct, fedorenko2016language}, although the causal manipulations are typically not as precise as those that can be accomplished in artificial systems.

Both encoders and decoders can be used to guide causal interventions. An encoder can discover ANN units that are strongly driven by certain features, which can then be ablated or altered to test causality. In contrast, if the information is distributed broadly throughout the system, we can use a decoder to discover the neural correlates of $Y$ and then alter them within the entire layer \citep{elazar2021amnesic}. Neither approach places theoretical constraints on desired probe expressivity\footnote{Although it can be speculated that simpler probes are more likely to identify causally relevant features, this has not been shown systematically.}.
One limitation of this approach is that, even if we establish that a given feature set plays a causal role in ANN's performance, we still cannot definitively establish the mechanisms that process these features. 
For instance, if we remove part of speech information from an early ANN layer, its performance will be impaired even if the actual part of speech processing happens several layers later. 

\subsection{Determining feature usability}

One question that can provide more mechanistic insight into the system of interest is whether the feature correlates can be easily extracted and used by downstream computational units. This question is best answered with a decoder that mimics such extraction \citep{kriegeskorte2019interpreting}. To ensure usability, many neuroscientists have restricted the space of possible mappings $F$ to be linear, arguing that this is the most faithful representation of readout by a downstream region/layer \citep[e.g., ][]{kriegeskorte2011pattern, naselaris2011encoding, kamitani2005decoding}. Although this is a reasonable proxy for usability in many research scenarios, the linear readout assumption might not always hold. In neuroscience, many dendritic computations are nonlinear, meaning that feature usability is much higher than that emulated with a linear probe. In NLP, the vast majority of layer-to-layer transformations are also nonlinear, in which case a linear probe is not a faithful representation of a downstream process. Thus, studies that aim to determine usability should be much more precise in articulating their reasons for placing specific constraints on $F$. 

An alternative to explicitly restricting probe expressivity is to use a minimum description length approach to determine the complexity of a particular mapping \citep{voita2020information}. Instead of answering a binary question ``can these features be extracted?'', the complexity estimate might help answer ``how easy are these features to extract?". Overall, the goal of determining feature usability brings us one step closer to a mechanistic understanding of a system of interest.

\subsection{Discovering emergent structure}
Finally, one may ask whether the system has developed a structured way of representing features of interest. Here, we care not only about the existence of a mapping $F$, but also about the exact form of this mapping.
Neuroscientists have referred to emergent structure in $X$ as its ``representational geometry'', developing multiple methods to discover and characterize it \citep[see, e.g., ][ for an overview]{kriegeskorte2019peeling, chung2021neural}. In ML, many probing studies have similarly aimed to elucidate the structure of feature representations within ANN layers. For instance, one of the first ANN probing studies showed that object representations gradually become more linearly separable in deeper layers  of object recognition ANNs \citep{alain2016understanding}.
More recently, in NLP, \citet{hewitt2019structural} used a method they term a \textit{structural probe} to discover connections between syntactic tree structure and the representational geometry of self-supervised language ANNs.
Both of these studies used decoding probes, but an encoder can also yield structural insights into the system.

Probe expressivity is of particular importance when investigating structural relations between features: the functional form of $F$ is the hypothesis made about representational structure.
In this case, the greater the simplicity of the functional form of $F$, the more interesting the finding.
The emergence of simple structure (e.g., linear separability) in highly nonlinear networks provides valuable insight into $X$. It can help identify biases, be used as a pedagogical tool in building approximate intuitions about model properties, and formulate fundamental principles of information processing in task-optimized systems.

\section{Conclusion}

As the ML community performs probing studies to understand increasingly complex ANNs, it encounters experimental design questions that may be addressed (at least to some extent) by insights from the longer tradition of probing in neuroscience.
We have argued that two design choices in probing -- direction and expressivity -- can be determined by a researcher's choice of one of neuroscience-inspired goals that underlie probing. We have further highlighted that only some of these research goals involve positing questions about the mechanisms underlying information processing in a system of interest. Overall, a clear formulation of research goals is essential for designing informative probes for future studies.

\section*{Acknowledgments}

This paper was inspired by the Generative Adversarial Collaboration (GAC) initiative organized by the Computational Cognitive Conference board. We thank the ``Linear Models'' team members --- Martin Schrimpf, Stefano Anzellotti, Evelina Fedorenko, and Leyla Isik, --- as well as workshop participants for their valuable insights and GAC organizers for their continued support. 
JH was supported by an NSF Graduate Research Fellowship under grant number DGE-1655618. NZ was supported by a BCS Fellowship in Computation.

\bibliography{iclr2021_conference}
\bibliographystyle{iclr2021_conference}

\end{document}